\documentclass[letterpaper]{article}
\usepackage{aaai}
\usepackage{times}
\usepackage{helvet}
\usepackage{courier}

\usepackage{booktabs,chemformula}
\usepackage{amsmath}
\usepackage{graphicx}

\begin{document}
%
\title{Facilitate the Parametric Dimension Reduction by Gradient Clipping}

\author{Chien-Hsun Lai \\
National Chiao Tung University \\ Hsinchu, Taiwan \\
jxcode.tw@gmail.com \\
\And
Yu-Shuen Wang\\
National Chiao Tung University \\
Hsinchu, Taiwan \\
yushuen@cs.nctu.edu.tw
}

\maketitle
\begin{abstract}
\begin{quote}
We extend a well-known dimension reduction method, t-distributed stochastic neighbor embedding (t-SNE), from non-parametric to parametric by training neural networks. The main advantage of a parametric technique is the generalization of handling new data, which is particularly beneficial for streaming data exploration. However, training a neural network to optimize the t-SNE objective function frequently fails. Previous methods overcome this problem by pre-training and then fine-tuning the network. We found that the training failure comes from the gradient exploding problem, which occurs when data points distant in high-dimensional space are projected to nearby embedding positions. Accordingly, we applied the gradient clipping method to solve the problem. Since the networks are trained by directly optimizing the t-SNE objective function, our method achieves an embedding quality that is compatible with the non-parametric t-SNE while enjoying the ability of generalization. Due to mini-batch network training, our parametric dimension reduction method is highly efficient. We further extended other non-parametric state-of-the-art approaches, such as LargeVis and UMAP, to the parametric versions. Experiment results demonstrate the feasibility of our method. Considering its practicability, we will soon release the codes for public use.

\end{quote}
\end{abstract}

\section{Introduction}
Dimension reduction (DR) techniques are widely utilized to facilitate data exploration and visual analysis. The goal is to project data from high-dimensional space to low-dimensional embedding space, while retaining either global or local data attributes. Among the techniques, t-distributed stochastic neighbor embedding (t-SNE) \cite{vandermaaten2008visualizing:tsne} is considered a classical method, which attempts to maximize the probability of nearby/distinct points in high-dimensional space to be nearby/distinct in embedding space. Subsequently, several extensions were presented to improve the performance of t-SNE because it has to update the conditional probability of each data point in the embedding space in each iteration.


T-SNE is a non-parametric DR method. The advantage of a non-parametric method is high flexibility when determining the positions of data points in low-dimensional embedding space. Without the parametric constraint, it can effectively optimize the objective function and obtain high- quality results \cite{journals/ijon/GisbrechtSH15:kernel-tsne}. On the other hand, non-parametric methods lack generalization, and are unable to apply the transformation obtained from the given dataset to reduce the dimensionality of new data points. To handle new data points, the methods have to merge the old and the new data sets, and then recompute the data positions in the embedding space. Since the goal of DR is retaining only data attributes, which is relative, the results in consecutive runs could differ by a rigid transformation. Users may not be able to compare the results quickly in the runs and must pay more attention when they study online or streaming data by using non-parametric methods.

We train a neural network to reduce data dimensionality by optimizing the t-SNE objective function. The network can be considered as a function, and is generalized to handle new data points. Although this idea is not new, none of the previous methods learn the function by directly optimizing the t-SNE objectives. They either require a network pre-training step by the restricted Boltzmann machine \cite{journals/jmlr/Maaten09:parametrictsne} or intermediate results obtained from the traditional non-parametric t-SNE \cite{journals/ijon/GisbrechtSH15:kernel-tsne} when learning the parametric mapping. Moreover, experiment results revealed that these parametric methods are not as good as those of non-parametric methods. In this study, we point out that the exploding gradient problem fails the network training when data points distant in high-dimensional space are projected to nearby embedding positions. We thus apply the gradient clipping to solve the problem. Because our network is trained from scratch by directly optimizing the t-SNE objective function, it greatly improves the quality of DR results compared to previous parametric methods. Furthermore, we adopt the commonly used stochastic gradient descent method to train the network, which considers only a batch of data when updating the network parameters in each iteration. The computation complexity of updating conditional probabilities of data points in the embedding space is reduced, and the system performance increases. 

To evaluate the effectiveness of our DR network, we compared the k-nearest neighbor (KNN) accuracy and trustworthiness of the results transformed by our network and previous methods. We also compared the performance of the methods in terms of the size of datasets. Experiment results show that our parametric method is compatible, or even outperforms, non-parametric methods, such as t-SNE, LargeVis, and UMAP, in terms of quality and performance, making our approach a highly practical DR solution. We also conducted several experiments to address network capacity, network architecture and out-of-sample generalization to evaluate its feasibility. We summarize our contributions as follows:

\begin{itemize}
    \item We present a parametric DR method that achieves not only generalization but also compatible quality and performance compared to non-parametric methods.
    
    \item We found that exploding gradients fail the training of t-SNE DR networks, explain the cause of the problem, and apply gradient clipping to facilitate network training.
    
    \item We conducted several experiments to evaluate the presented DR network in terms of network capacity, architecture, data sizes, and generalization ability.
\end{itemize}

\section{Related Works}

\subsection{Dimension Reduction}
Dimension reduction algorithms project high-dimensional data to low-dimensional embedding space, which allows users to visually analyze data structures and identify outliers. Because the dimensionality of data is reduced, information loss and distortions are inevitable. The DR algorithms aim to retain data attributes during dimension reduction. Among the linear methods, principal component analysis (PCA) \cite{Smith2002:PCA} seeks to maximize the variance of data after they are projected to low-dimensional space. Linear discriminant analysis (LDA) \cite{Izenman2008:LDA} extends the PCA by incorporating data labels, which aims to optimally separate data points of different labels. Non-linear methods, such as multi-dimensional scaling (MDS) \cite{BorgGroenen2005:MDS1,kruskal1964:MDS2,journals/csur/SaeedNHB18:MDS3}, isometric feature mapping (IsoMap) \cite{tenenbaum_global_2000:isomap}, self-organizing map (SOM) \cite{VanHulle2012:SOM}, locally linear embedding (LLE) \cite{roweis2000nonlinear}, maximum variance unfolding (MVU) \cite{weinberger2006introduction}, and Laplacian eigenmaps \cite{belkin2003laplacian}, were presented to preserve the relative distances of data points in high- and low-dimensional spaces to be similar.



Stochastic neighbor embedding (SNE) \cite{hinton2003stochastic} models the relationship between data points by a conditional probability, rather than distances. By minimizing the KL divergence of data distributions in high- and low- dimensional spaces, it faithfully preserves the local structure of data. Subsequently, t-SNE \cite{vandermaaten2008visualizing:tsne} extends the SNE by symmetrizing the probability distribution and employing a heavy-tailed student t-distribution to compute the joint probability of data points in the embedding space. To preserve other attributes of high-dimensional data, such as global data distances, \citeauthor{journals/corr/abs-1811-01247:sne-fdivergence} replaced the KL divergence with f-divergence metrics for different types of structure discovery. LargeVis \cite{conf/www/TangLZM16:largevis} uses a similar strategy to t-SNE but eliminates the need for normalization in the embedding space. In other words, it optimizes the objective function using stochastic gradient descent and is scalable to large datasets. UMAP \cite{mcinnes2018uniform:umap} utilizes the language of algebraic topology to preserve the local distance structure. It achieves comparable quality to t-SNE and LargeVis, while being able to retain the global structure of data and consume lower computation cost.

\subsection{Parametric Extensions of t-SNE}
T-SNE and its extensions have proven to work well on many real-world datasets. One drawback, however, is the lack of an explicit mapping function to handle unseen data. Consequently, several methods were presented to extend t-SNE from non-parametric to parametric at the cost of lower embedding quality, which is resultant from the non-flexibility of a parametric form. Parametric t-SNE \cite{journals/jmlr/Maaten09:parametrictsne} uses a stack of restricted Boltzmann machine to pre-train a feed-forward network, and then fine-tunes the network using the t-SNE objective function. The authors claimed that the pre-training was needed because of the complex parameter interactions. Otherwise, the network would be stuck in a bad local minimum if it is updated by backpropagation. Subsequently, dt-SEE \cite{conf/ijcai/MinGS17:dtsee} extends the parametric t-SNE with exemplars in high-dimensional space to avoid pairwise distance calculation. Besides the neural network-based parameterization methods, \citeauthor{journals/neco/BunteBH12:local-linear} combined global linear mapping and piece-wise local linear mapping to project data points. They divided the high-dimensional data into several receptive fields, and mapped data in each field by a linear transformation. The kernel t-SNE \cite{conf/esann/GisbrechtLMH12:kernel,journals/ijon/GisbrechtSH15:kernel-tsne} models the paired input and the embedding produced by the traditional t-SNE by Gaussians. To reduce the dimensionality of unseen data, it computes coefficients for linearly interpolating the kernels in the embedding space. 

Previous parametric methods, except piece-wise local linear mapping \cite{journals/neco/BunteBH12:local-linear}, require either intermediate results or pre-training to optimize the t-SNE objective function. Although the piece-wise local linear mapping adopts the direct optimization strategy, the linear transformation inherits a strong regularization. As a consequence, none of the previous parametric methods achieve competitive performance compared to non-parametric methods. In contrast, our neural network based algorithm, which is highly nonlinear and is trained by direct optimization of the loss, performs as well as non-parametric methods while simultaneously enjoying the ability of generalization.



\section{Background}
We first describe the objective of the non-parametric t-SNE and then extend the method to a parametric version. Given a data set $X = \{x_i \in R^{d}\} $ in high-dimensional space and the embedding set $Y = \{y_i \in R^{s}\}$ that contains the corresponding points in low-dimensional space. The DR method attempts to find a mapping $C$ that can minimize a predefined loss function
\begin{equation}
    \operatorname*{argmin}_Y C(X, Y)
    \label{eq:DR}
\end{equation}
to retain data attributes. In each iteration, the algorithm calculates the gradient of the loss with respect to the embedding $\frac{\partial C}{\partial y_i}$ and moves each $y_i$ to the desirable position. For simplicity, we use the term \emph{embedding} to denote data points in low-dimensional space in later sections.

T-SNE is an extension of the SNE. Specifically, SNE attempts to maintain points that are nearby/distinct in high-dimensional space to be nearby/distinct in low-dimensional space. To implement this idea, SNE models the similarity of data points $x_j$ to $x_i$ by the conditional probability $p_{j|i}$, which is a Gaussian centered at $x_i$. This can be expressed as:
\begin{equation}
    p_{j|i} = \frac{exp(-||x_i-x_j||^2 / 2\sigma_i^2)}{\sum_{k!=i} exp(-||x_i-x_k||^2 / 2\sigma_i^2)},
    \label{eq:tsne-cond-prob}
\end{equation}
where $\sigma_i$ is the variance determined by the neighbor perplexity. Similarly, for the corresponding points $y_j$ and $y_i$ in low-dimensional space, we can compute a conditional probability $q_{j|i}$. The goal is to make the two conditional probabilities $p_{j|i}$ and $q_{j|i}$ to be equal. Therefore, SNE minimizes the mapping function:
\begin{equation}
    C = \sum_i{KL(P_i||Q_i)} = \sum_i{\sum_j{p_{j|i}log\frac{p_{j|i}}{q_{j|i}}}},
\label{eq:sne-kld}
\end{equation}
where $KL$ is the Kullback-Leibler (KL) divergence, $P_i = \{p_{j|i}\}$ and $Q_i = \{q_{j|i}\}$, to maintain the conditional probabilities of all data points.

T-SNE improves SNE in two ways. First, it symmetries the conditional probability to a joint probability by setting $p_{ij} = \frac{p_{j|i}+p_{i|j}}{2n}$. The main advantage of the symmetry is the simple computation of its gradient. Second, to alleviate the crowding problem, t-SNE employs a heavy-tailed student t-distribution to compute a weight $w_{ji}$ between $y_j$ and $y_i$. It then normalizes the weight to obtain the joint probability as follows:
\begin{equation}
    q_{ij} = \frac{w_{ij}}{\sum_k\sum_l w_{kl}}, \;w_{ij}=\frac{1}{1+||y_i-y_j||_2^2}.
\label{eq:tsne_Q}
\end{equation}
Let $P$ and $Q$ be the joint distributions of $p_{ij}$ and $q_{ij}$, respectively. The KL-divergence of the conditional probability is now changed to measure the divergence of the joint probability
\begin{equation}
    C = KL(P||Q) = \sum_i{\sum_j{p_{ij}log\frac{p_{ij}}{q_{ij}}}} \quad.
\label{eq:tsne-kld}
\end{equation}
Then, one can derive the gradient concerning $y_i$ as
\begin{equation}
    {\partial C \over \partial y_i} = 4\sum_j {(p_{ij}-q_{ij})(y_i-y_j) \over 1+||y_i-y_j||_2^2}
\label{eq:tsne-grad}
\end{equation}
to update point positions in low-dimensional space.

\section{Neural Network as a Parametric Dimension Reduction Method}
We extend the non-parametric t-SNE to a parametric t-SNE by training a deep neural network. The network $f_\theta(x_i)$ maps each input data point $x_i$ to the embedding $y_i$, where $\theta$ indicates the parameters of $f$. Accordingly, we rewrite Eq. \ref{eq:DR} as
\begin{equation}
    \operatorname*{argmin}_{\theta} C(X, f_\theta(x_i)), \quad \forall x_i \in X.
\end{equation}
Because our goal is to compute the function $f$, the unknowns are the network parameters $\theta$. The gradient then becomes
\begin{equation}
    \frac{\partial C}{\partial \theta} = \frac{\partial C}{\partial f_\theta(x_i)} \frac{\partial f_\theta(x_i)}{\partial \theta}, \quad \forall x_i \in X \label{eq:parametric_dr_gradient}
\end{equation}
for updating the network. By substituting $\frac{\partial C}{\partial f_\theta(x_i)}$ in Eq. \ref{eq:parametric_dr_gradient} with Eq. \ref{eq:tsne-grad}, we obtain the following:
\begin{align}
     \frac{\partial C}{\partial \theta} = 4 \sum_i \sum_j {(p_{ij}-q_{ij}) (f_\theta(x_i)-f_\theta(x_j)) \over 1+||f_\theta(x_i))-f_\theta(x_j)||_2^2} \frac{\partial f_\theta(x_i)}{\partial \theta}.
     \label{eq:ours-tsne-loss}
\end{align}

Optimizing the network $f$ using Eq. \ref{eq:ours-tsne-loss} demands the computation of each joint probability $q_{ij}$ and $f_\theta(x_i)-f_\theta(x_j)$ whenever the network is updated. The computation complexity is $O(N^2)$, where $N$ is the number of data points in the whole dataset. Note that neural networks are trained by using the stochastic gradient descent method, and in each iteration, only a batch of data points are sampled from the data set and used to update the network parameters. Therefore, we compute the joint probability $q_{ij}$ based on only the sampled data points in our implementation. The complexity is then reduced from $O(N^2)$ to $O(n^2)$ in each iteration, where $n$ is the batch size. 



\begin{figure}
    \centering
    \includegraphics[width=\columnwidth]{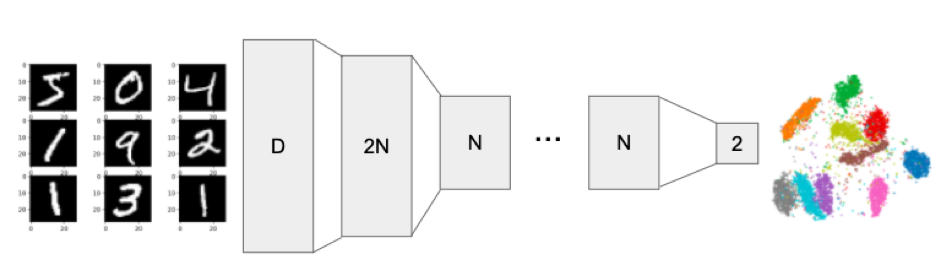}
    \caption{Overview of the encoder network. D is the input dimension, and N is the number of hidden units. By default, N = 256, and there are three layers.}
    \label{fig:network}
\end{figure}

Figure \ref{fig:network} shows the default neural network architecture used in our experiments. Without a loss of generality, the network by default is composed of fully connected layers. The activation function in each layer is Leaky ReLU \cite{Maas2013RectifierNI:relu}. We also carried out experiments to test the performance of convolution neural networks because of the experiments on the image datasets. The results are shown in the later section.


\subsection{Gradient Clipping}
\begin{figure}[ht]
    \centering
    \includegraphics[width=\columnwidth]{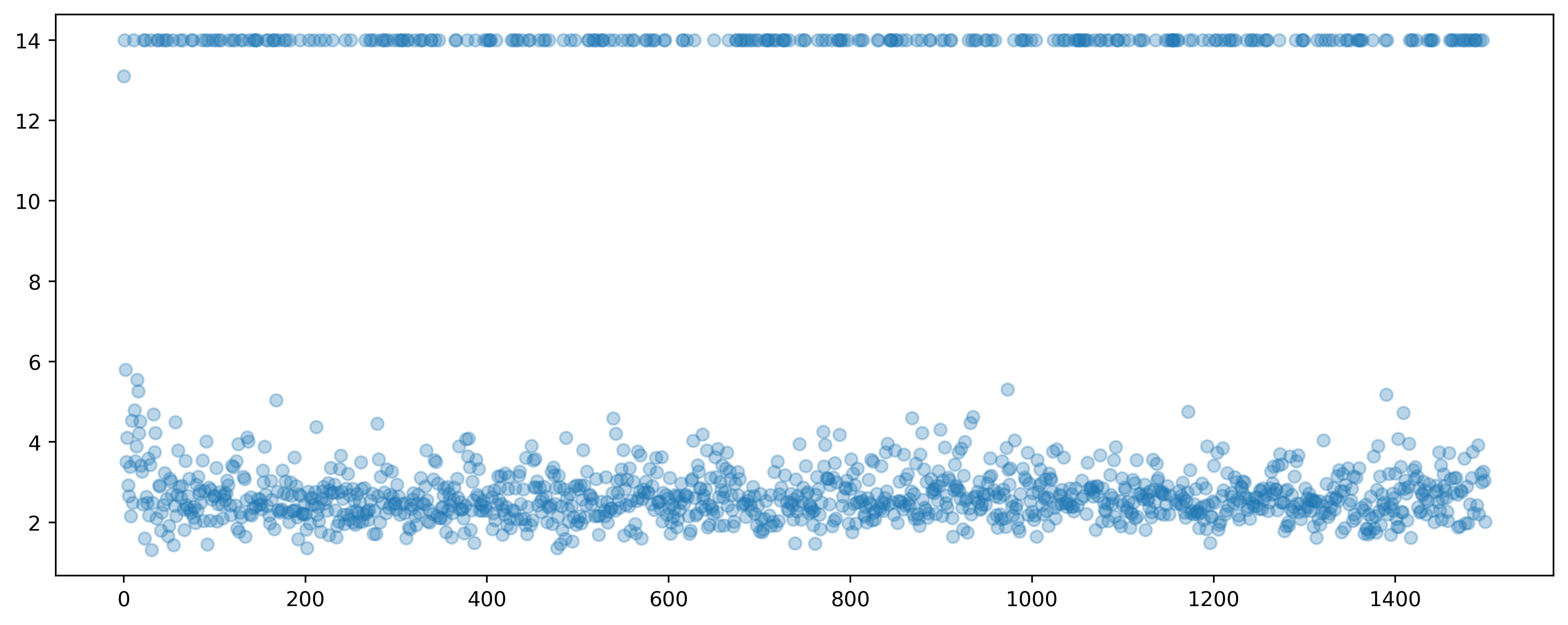}
    \caption{ The gradient magnitude is clipped to $10^{14}$ to avoid the gradient exploding problem. We display the clipped maximum gradients (Y-axis) in each iteration (X-axis). The Y-axis is log scaled.}
    \label{fig:norm}
\end{figure}

\begin{figure}[ht]
    \centering
    \includegraphics[width=\columnwidth]{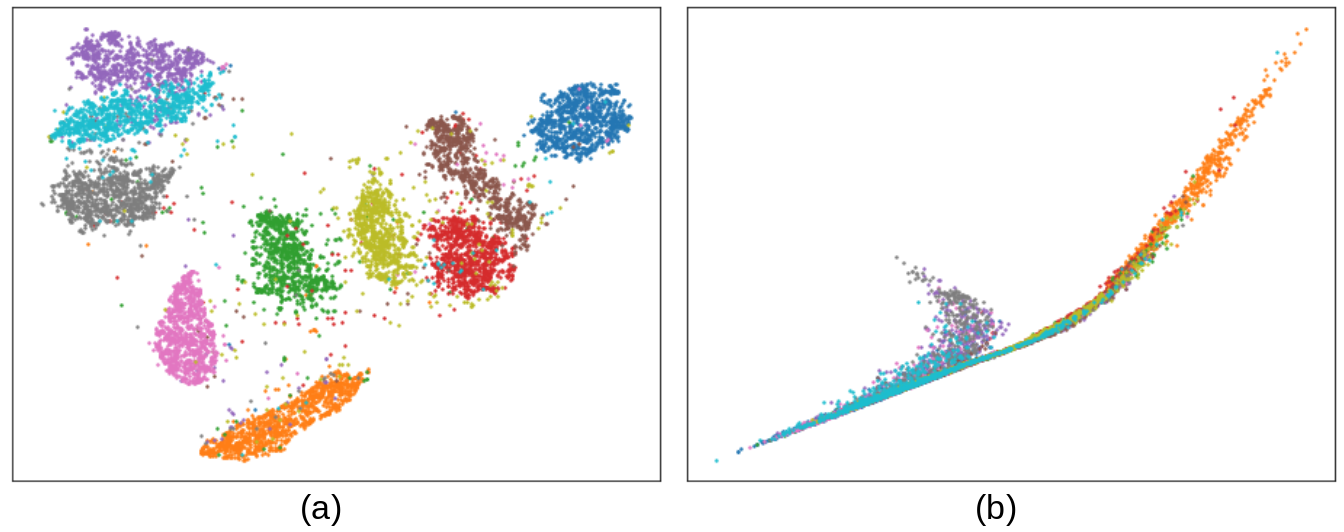}
    \caption{Experiment on the MNIST dataset with (a) and without (b) the gradient clipping step. As can be seen, 10 classes of hand written numbers are mixed together if the gradients are not clipped.}
    \label{fig:clip_comparison}
\end{figure}

The loss of t-SNE involves the computation of $log({p_{ij} \over q_{ij}})$ (Eq. \ref{eq:tsne-kld}), which could be close to zero. As a result, the gradient that is used to update the network parameters will be multiplied by a large value of $log({q_{ij} \over p_{ij}})$ and fails the network training. The gradient exploding problem occurs when data points distant in high-dimensional space are projected to nearby positions in low-dimensional space. This situation is commonly seen in DR because the network parameters are randomly initialized. In addition, the capacity of representing data attributes has been reduced, and distortions in some cases are difficult to avoid. Figure \ref{fig:norm} (a) indicates that gradients with huge magnitudes constantly appear throughout the network training. To solve this problem, we clip the gradients before they are used in the backpropagation step. In our experiments, we set the clipping threshold to $10^{14}$. We also clip the gradient of each layer within the magnitude of $10^4$ for the stable network training. Figure \ref{fig:clip_comparison} shows the comparison of the DR results with and without gradient clipping.



\section{Training Details}
We have implemented the presented DR method using PyTorch \cite{paszke2019pytorch:pytorch}. We used the Xavier initialization \cite{pmlr-v9-glorot10a:xavier-init} to initialize the network parameters, set the batch size to 1024, and set the learning rate to $10^{-3}$. To update network parameters, we used the RMSProp optimizer \cite{hinton2012neural:rmsprop}. The widely used Adam optimizer \cite{Adam} was not adopted because it incorporates the momentum to the optimization process, which would lead to unstable network training because large gradients in consecutive steps may disturb each other. In our experiments, we found that both the SGD and RMSProp optimizers work equally well, but the RMSProp converges faster. 
Moreover, we used the early exaggeration trick the same as the original t-SNE method, in the first 250 iterations for a faster convergence. Finally, and most importantly, we clipped the gradients if their magnitudes were larger than $10^{14}$.



\section{Experiments and Results}
We compared our parametric t-SNE, called \emph{nn-tsne}, with state-of-the-art parametric and non-parametric methods to demonstrate its effectiveness. Considering the great success of LargeVis \cite{conf/www/TangLZM16:largevis} and UMAP \cite{mcinnes2018uniform:umap} in DR, we also extended the non-parametric LargeVis and UMAP to the parametric versions similarly by training a neural network and clipping extremely large gradients. These two methods were denoted as \emph{nn-largevis} and \emph{nn-umap}, respectively (see Appendix for details). Different from \emph{nn-tsne}, \emph{nn-largevis} and \emph{nn-umap} were trained by optimizing their own objective functions. Furthermore, we followed the negative data sampling strategies presented in the original paper since relations could exist between the objective function and the sampling strategy. From the experiments, we also verified that the original sampling strategies performed better than the mini-batch sampling.

To achieve an objective evaluation, we project high-dimensional data to 2D planar space and use the KNN classifier to identify object categories. The KNN accuracy is high if the embedding preserves the relative data positions well. We also use trustworthiness \cite{journals/ijon/LeeV09:trustworthiness} as a metric for assessing the embedding quality, which expresses to what extent the local structure is preserved. The values of trustworthiness range from 0 to 1
, in which a higher value indicates better quality. Note that we repeated all of the experiments five times. All of the numbers shown in the tables were averages of five runs. In addition, we visualize the 2D embeddings for subjective evaluations. Readers can observe whether data samples in the same category were close to each other and whether gaps between samples in different categories were clear in order to determine the embedding quality.




\begin{table}[t]
  \centering
  \caption{KNN accuracy comparison between classic non-parametric dimension reduction, supervised training (sup), and our method (ours) under the same type of loss function.}
  \begin{tabular}{@{}cccc@{}}
    \toprule
    Method  & Coil-20 & Fashion-MNIST & MNIST \\
    \midrule
    AutoEncoder   & 89.7\% & 68.0\% & 82.5\% \\
    \midrule
    t-SNE         & 99.4\% & 79.6\% & 94.3\% \\
    t-SNE(ours)   & 97.1\% & 78.3\% & 93.3\% \\
    \midrule
    LargeVis      & 97.2\% & 73.6\% & 91.9\% \\
    LargeVis(ours)& 92.7\% & 67.5\% & 89.6\% \\
    \midrule
    UMAP          & 91.0\% & 71.9\% & 89.3\% \\
    UMAP(ours)    & 88.8\% & 69.0\% & 89.1\% \\
    \bottomrule
  \end{tabular}
  \label{table:knn_accuracy}
\end{table}

\begin{table}[h]
  \centering
  \caption{Trustworthiness for each method.}
  \begin{tabular}{@{}cccc@{}}
    \toprule
    Method  & Coil-20 & Fashion-MNIST & MNIST \\
    \midrule
    AutoEncoder   & 0.987 & 0.977 & 0.974 \\
    \midrule
    t-SNE         & 0.998 & 0.993 & 0.993 \\
    t-SNE(ours)   & 0.993 & 0.989 & 0.964 \\
    \midrule
    LargeVis      & 0.997 & 0.986 & 0.979 \\
    LargeVis(ours)& 0.983 & 0.964 & 0.946 \\
    \midrule
    UMAP          & 0.993 & 0.980 & 0.962 \\
    UMAP(ours)    & 0.987 & 0.974 & 0.958 \\
    \bottomrule
  \end{tabular}
  \label{table:trustworthiness}
\end{table}

\subsection{Comparison with Non-parametric Methods}
We first compared our \emph{nn-tsne}, \emph{nn-largevis}, and \emph{nn-umap} to non-parametric state-of-the-art approaches, including the traditional t-SNE \cite{vandermaaten2008visualizing:tsne}, LargeVis \cite{conf/www/TangLZM16:largevis}, and UMAP \cite{mcinnes2018uniform:umap}, to demonstrate that the parametric extensions were competitive with non-parametric methods in terms of quality. In the comparison, we obtained the results of t-SNE from openTSNE \cite{Policar731877:opentsne}, and the results of LargeVis and UMAP from the authors' released codes. The comparison was conducted on Coil-20 \cite{nane1996columbia:coil20}, MNIST \cite{dataset:mnist}, and Fashion-MNIST \cite{xiao2017fashionmnist:fmnist} datasets. Intuitively, the non-parametric dimension reductions have the highest degree of flexibility in the embedding and should have the best quality \cite{journals/ijon/GisbrechtSH15:kernel-tsne}. The results in Tables \ref{table:knn_accuracy} and \ref{table:trustworthiness} verified this assertion: the traditional t-SNE outperforms our \emph{nn-tsne}, and \emph{nn-largevis} and \emph{nn-umap} were less competitive with their non-parametric versions in terms of embedding quality. This phenomenon was reasonable due to the regularization constraint. However, the gap between the parametric and our non-parametric extensions was small. Note that, overall, \emph{nn-tsne} achieved better embedding quality than the non-parametric LargeVis and UMAP.



\begin{figure}[ht]
    \centering
    \includegraphics[width=\columnwidth]{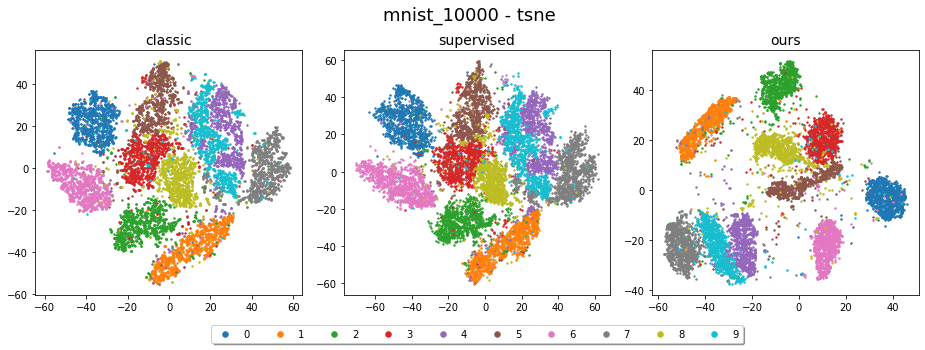}
    \caption{(a) Embedding produced by t-SNE. (b) Supervisory training an encoder using the embedding in (a). (c) Embedding produced by our method (t-SNE loss).}
    \label{fig:supervised_vs_ours}
\end{figure}

It is worth noting that the regularization constraint from a parametric form does not always diminish the quality. Figure \ref{fig:supervised_vs_ours} (a) presents an example of this. Although the traditional t-SNE achieved higher KNN accuracy and trustworthiness than our \emph{tsne-nn} did, the high degree of flexibility could map data points in the same category to several fragmented clusters. The experiment on the MNIST dataset showed that the traditional t-SNE frequently mixed the digits 4 (purple) and 9 (light blue) in the embedding. The results would lead to misinterpretations during data exploration. 



\subsection{Comparison with Parametric Methods}
\begin{figure}[ht]
    \centering
    \includegraphics[width=\columnwidth]{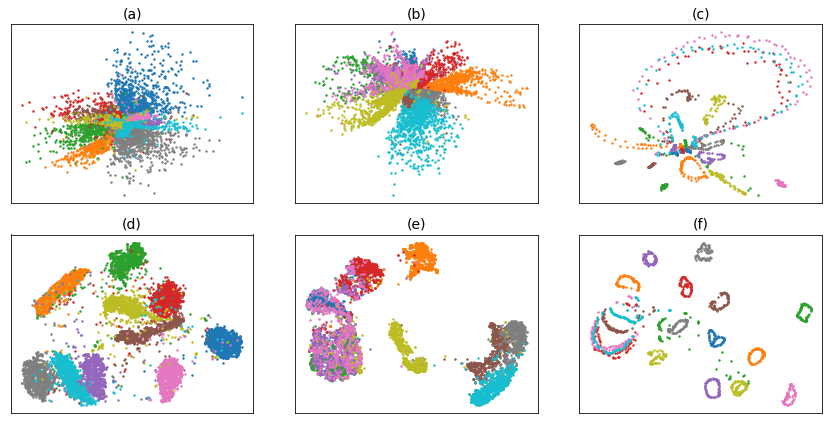}
    \caption{(a-c) Embeddings computed by the autoencoder on MNIST, Fashion MNIST, and Coil-20. (d-f) Embeddings learned by our method.}
    \label{fig:autoencoder_vs_ours}
\end{figure}

The autoencoder \cite{HintonSalakhutdinov2006b:autoencoder} has achieved great success in learning low-dimensional embedding of data in many applications. To achieve a fair comparison, we used the same encoder architecture. The decoder was symmetric to the encoder. Figure \ref{fig:autoencoder_vs_ours}, and Tables \ref{table:knn_accuracy} and \ref{table:trustworthiness}, show the results. As can be seen, the autoencoder projected data to a more scattered embedding. Data points of the different categories were not clearly separated. Therefore, it achieved lower KNN accuracy and trustworthiness compared to \emph{nn-tsne}. These results were reasonable because the objective of an autoencoder was reconstruction. Specifically, the embedding was learned indirectly by distinguishing data points during reconstruction. 




\begin{table}[h]
  \small
  \centering
  \caption{Training set KNN accuracy for each method. The results are taken from the kernel t-SNE paper \cite{journals/ijon/GisbrechtSH15:kernel-tsne}.}
  \begin{tabular}{@{}cccccc@{}}
    \toprule
    Dataset &  & Parametric & Kernel & Fisher kernel & nn-tsne \\
      & & t-SNE & t-SNE & t-SNE &  \\
    \midrule
    Letter & Train & 21.3\% & 84.1\% & 85.5\% & 94.3\% \\
           & Test  & 27.8\% & 80.1\% & 80.4\% & 79.7\% \\
    MNIST  & Train & 85.4\% & 90.7\% & 91.1\% & 93.4\% \\
           & Test & 62.5\% & 85.8\% & 86.3\% & 87.8\% \\
    Norb   & Train & 43.0\% & 88.2\% & 85.4\% & 97.0\% \\
           & Test & 38.5\% & 85.4\% & 85.6\% & 75.6\% \\
    USPS   & Train & 86.5\% & 90.5\% & 96.6\% & 96.8\% \\
           & Test & 58.6\% & 84.8\% & 87.4\% & 88.9\% \\
    \bottomrule
  \end{tabular}
  \label{table:cmp_previous}
\end{table}

\begin{figure}[ht]
    \centering
    \includegraphics[width=\columnwidth]{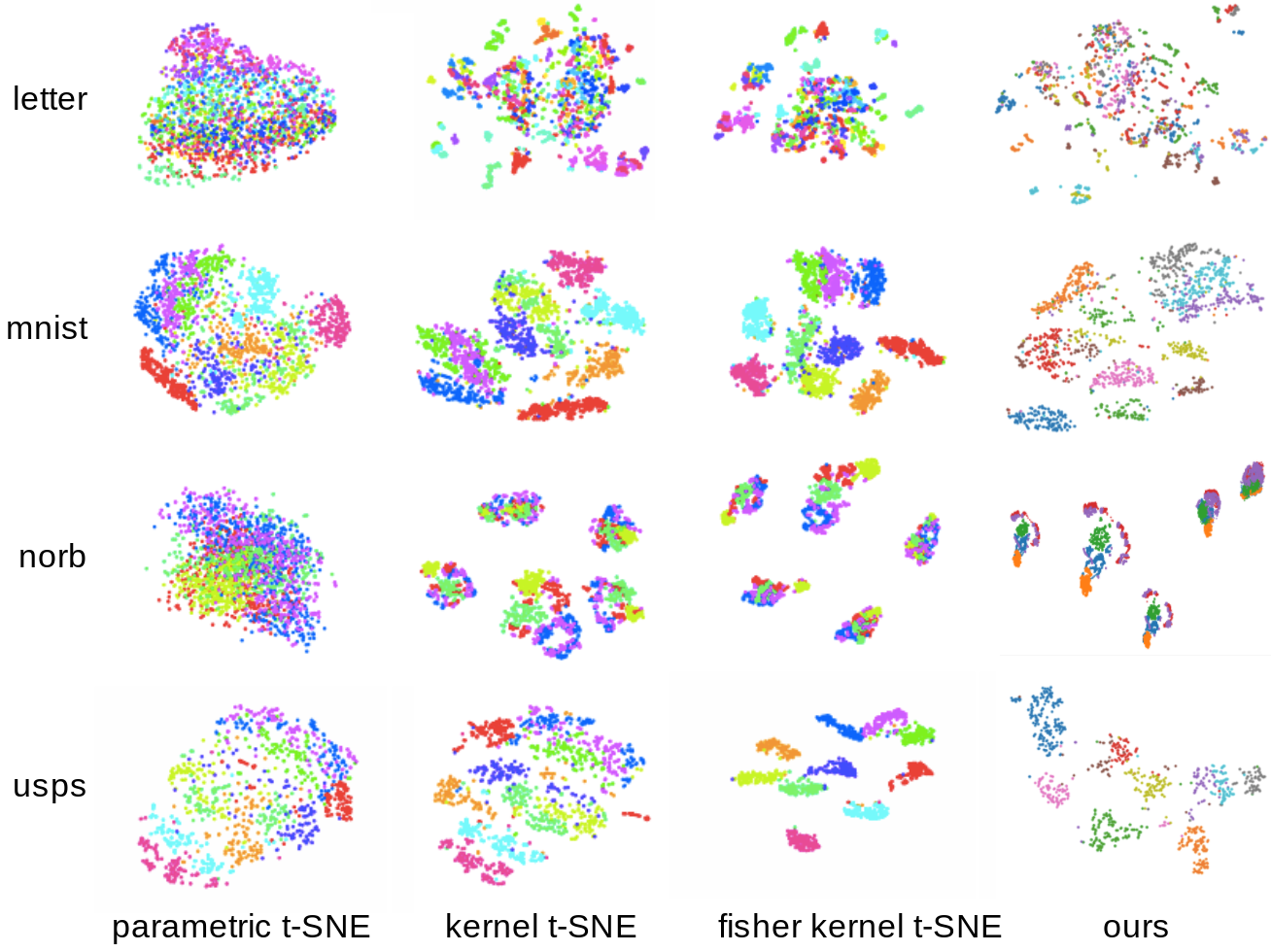}
    \caption{Comparison of our method and previous methods in various datasets. The first three columns were reported from a previous work on kernel t-SNE \cite{journals/ijon/GisbrechtSH15:kernel-tsne}.}
    \label{fig:2d_embedding_cmp}
\end{figure}

\begin{figure}[ht]
    \centering
    \includegraphics[width=\columnwidth]{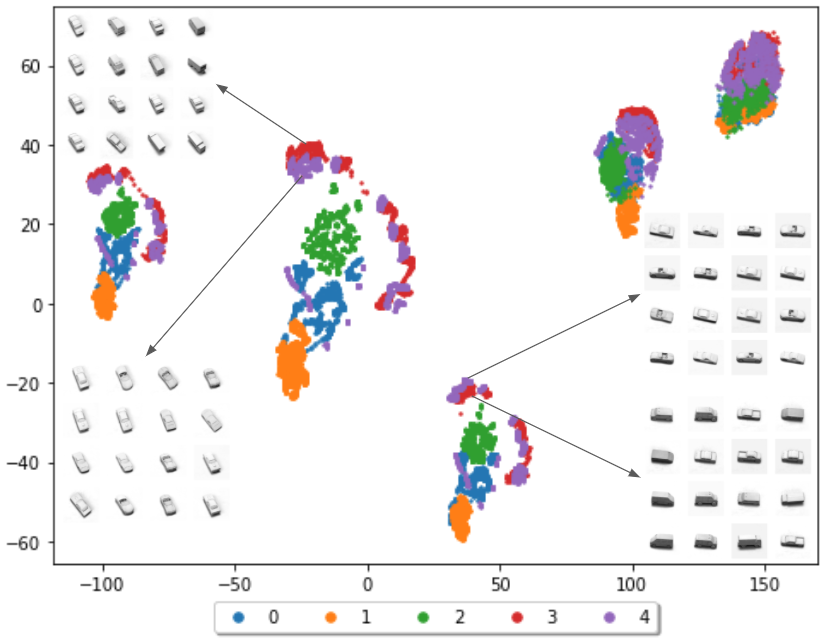}
    \caption{Embedding of the Norb dataset transformed by our \emph{nn-tsne}. The red and purple dots represent the truck and car classes, respectively. Dots in a cluster indicate the same orientation, and truck and car images are visually similar.}
    \label{fig:norb_trucks_cars}
\end{figure}

Finally, we compared our \emph{nn-tsne} to parametric t-SNE \cite{journals/jmlr/Maaten09:parametrictsne} and kernel t-SNE \cite{journals/ijon/GisbrechtSH15:kernel-tsne} by running the experiments in the same manner as those described in the work of kernel t-SNE. Specifically, the Letter \cite{dataset:letter}, MNIST \cite{dataset:mnist}, Norb \cite{dataset:norb}, and USPS \cite{dataset:usps} datasets were used, and KNN accuracy was computed for each embedding. Table \ref{table:cmp_previous} shows the results, in which the numbers of the previous methods were reported in the paper of kernel t-SNE. The Fisher kernel t-SNE was a variant of kernel t-SNE, which utilized auxiliary label information when computing the embedding. As can be seen, our \emph{nn-tsne} outperformed parametric t-SNE and kernel t-SNE by a large margin on all of the training datasets, but not on all of the testing sets. One also can observe a significant drop of KNN accuracy between the training and the testing Norb dataset. We surmised that the reason for this was the invalid distance measure of data points in high-dimensional space. Figure \ref{fig:norb_trucks_cars} shows two classes of images, which were labeled as trucks and cars, respectively. Although the two classes of images were different, the Euclidean distance between them was quite short. Accordingly, each cluster in the embedding space represents a viewpoint instead of a class. While the gaps between classes were quite miniscule, the significant decrease of KNN accuracy on the testing set was reasonable.


\subsection{Network Capacity}
\begin{figure}[t]
    \centering
    \includegraphics[width=\columnwidth]{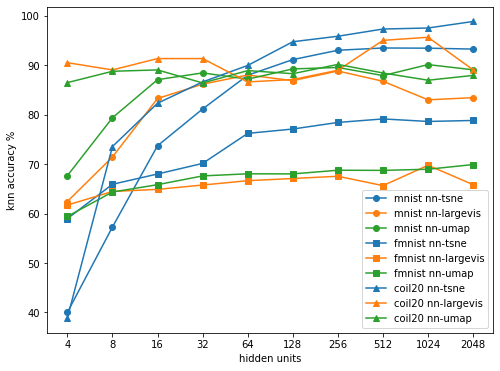}
    \caption{KNN accuracy to number of hidden units on MNIST (circle), Fashion MNIST (square), and Coil-20 (triangle), respectively.}
    \label{fig:network_capacity_knn}
\end{figure}

\begin{figure}[t]
    \centering
    \includegraphics[width=\columnwidth]{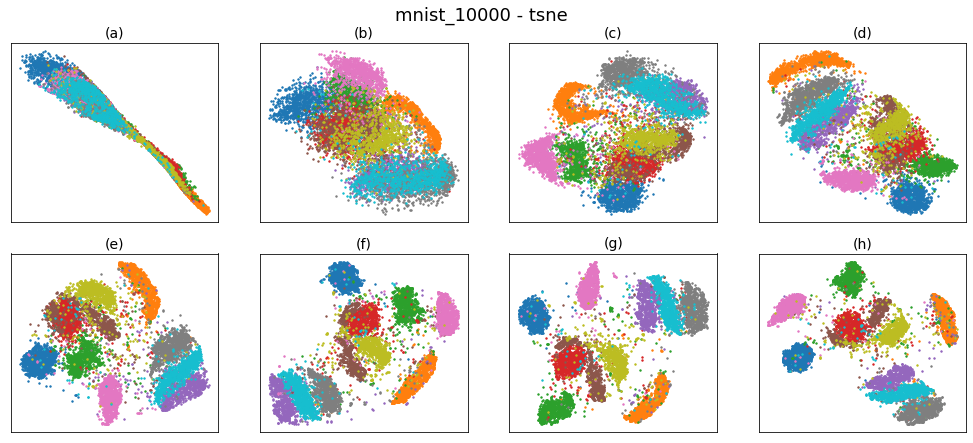}
    \caption{(a)~(h) results for different numbers of hidden units (4, 8, 16, …, 512) on the MNIST dataset with t-SNE loss.}
    \label{fig:network_capacity_2d}
\end{figure}

The encoder network was trained to reduce data dimensionality. We aimed to determine how network capacity affects embedding quality. Hence, we tested our \emph{nn-tsne}, \emph{nn-largevis}, and \emph{nn-umap} on the MNIST, Fashion MNIST, and Coil-20 datasets with different numbers of hidden units in the encoder. Figure \ref{fig:network_capacity_knn} presents the results. It is clear that the networks with larger capacities perform better, and KNN accuracy converges when the capacity reaches 512. Interestingly, the experiments revealed that \emph{nn-largevis} and \emph{nn-umap} were less influenced by the number of hidden layers, which could be used on devices with low computation power. We plotted the embeddings of the MNIST dataset in Figure \ref{fig:network_capacity_2d} for a visual comparison.


\subsection{Network Architecture}
Since we trained neural networks on several image datasets and evaluated their embedding qualities, we next compare the results transformed by the networks composed of fully-connected layers and convolutional layers. For this purpose, we additionally trained a convolutional neural network, in which the number of parameters is similar to those of a fully-connected network, for the evaluation. Table \ref{table:cmp_arch} shows that the embedding qualities of the two networks were similar, perhaps due to the same objective functions.

\begin{table}[h]
  \centering
  \caption{Statistics of fully connected and convolutional encoders.}
  \begin{tabular}{@{}cccc@{}}
    \toprule
    Methods & MNIST & Fashion-MNIST & Coil-20 \\
    \midrule
    nn-tsne(fc) & 93.4\% & 78.3\% & 97.1\% \\
    nn-tsne(cnn) & 95.0\% & 79.3\% & 96.5\% \\
    
    nn-largevis(fc) & 89.6\% & 67.4\% & 92.7\% \\
    nn-largevis(cnn) & 92.1\% & 63.6\% & 87.2\% \\
    
    nn-umap(fc)  & 89.0\% & 69.0\% & 88.7\% \\
    nn-umap(cnn) & 90.1\% & 68.6\% & 87.2\% \\
    \bottomrule
  \end{tabular}
  \label{table:cmp_arch}
\end{table}

\subsection{Batch Size}
\begin{figure}[ht]
    \centering
    \includegraphics[width=\columnwidth]{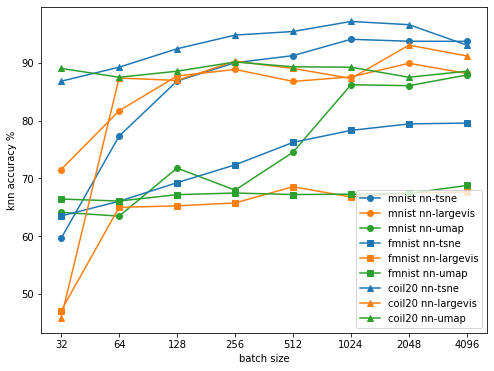}
    \caption{KNN accuracy increases as batch size increases because each mini-batch approximately represents the original dataset's distribution.}
    \label{fig:batch_size_cmp}
\end{figure}

It is the case that \emph{nn-tsne}, \emph{nn-largevis}, and \emph{nn-umap} all have to update the conditional probability of each data point in the embedding space in each iteration. To prevent high computation cost from considering the whole dataset, one can sample a batch of data points from the original dataset and use them to update the network. Since each mini-batch approximately represents the original dataset's distribution, the batch size should be sufficiently large to reduce uncertainty; whereas, a large batch size, increases the computation cost drastically. Figure \ref{fig:batch_size_cmp} shows the KNN accuracies of the dimension reduction results achieved by different batch sizes. The experiments suggest setting the batch size to 1024.

\begin{figure}[ht]
    \centering
    \includegraphics[width=\columnwidth]{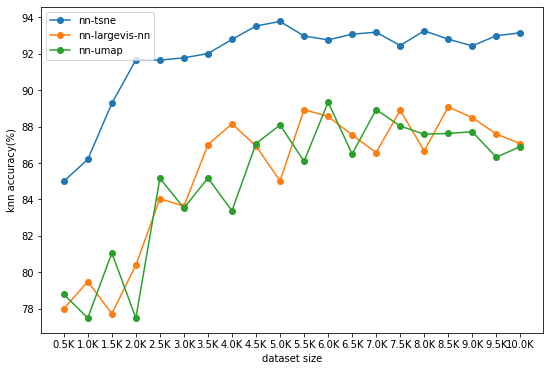}
    \caption{Experiment on the MNIST dataset by testing accuracy with respect to the size of the training set. The training set size ranged from 500 to 10,000.}
    \label{fig:generalization}
\end{figure}

\subsection{Out-of-Sample Generalization}

Since the main advantage of DR networks is generalization, we tested the networks by projecting unseen data and then measured the embedding quality. Specifically, we trained the networks on 0.5K to 10K images, which were randomly selected from the MNIST training set. The networks were then evaluated on the 10K MNIST testing set. Figure \ref{fig:generalization} shows that \emph{nn-tsne} had the best generalization ability, which could achieve 90\% KNN accuracy when it was trained on 2K images. \emph{nn-largevis} and \emph{nn-umap} could also generalize to the testing sets, with KNN accuracies of approximately 88\%.




\subsection{Run-time Comparison}
\begin{figure}[ht]
    \centering
    \includegraphics[width=\columnwidth]{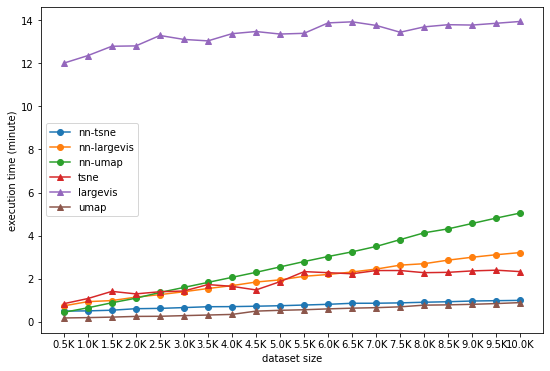}
    \caption{Timing statistics of the DR methods measured under different sizes of the MNIST dataset. The t-SNE and the LargeVis implementations were based on the CPU. The t-SNE implementation was a highly optimized version presented by \cite{linderman2019fast:tsne-acc2}.}
    \label{fig:runtime}
\end{figure}

Our \emph{nn-tsne} was competitive with previous non-parametric methods in terms of not only quality, but also computation performance. Figure \ref{fig:runtime} shows the timing statistics of the methods under various dataset sizes. All of the experiments were run on a server with two Intel(R) Xeon(R) CPU @ 2.20GHz, Nvidia Titan X GPU, and 12 GB of RAM. The MNIST dataset was used in the experiments and the programs stopped when the results converged. It is worth noting that a fair run-time comparison among the methods was difficult because of several issues. First, the implementations of the LargeVis and t-SNE were based on CPU, while the others were based on GPU. Second, the running time could be affected by hyperparameters such as the learning rate and the initial guess. Accordingly, from the results, we could only conclude that our \emph{tsne-nn} was as efficient as previous methods and was scalable to large datasets.


\subsection{Limitations}
Our parametric DR method is competitive with non-parametric state-of-the-art methods while simultaneously enjoying the ability of generalization. However, the embedding results depend heavily on the distance measure of data points in high-dimensional space. Specifically, it must utilize human labels or collaborate with other self-training techniques to generate semantic embeddings. Otherwise, the visual analysis could be meaningless, as illustrated in Figure \ref{fig:norb_trucks_cars}. Moreover, although we found several interesting phenomena in our experiments, such as the non-sensitivity of hidden units by the \emph{nn-largevis} and \emph{nn-umap}, at this moment, we could not analyze the phenomena from a theoretical perspective. Considering the success achieved in practice, we plan to investigate the fundamental theory and elucidate how neural networks reduce the dimensionality of data in the future.



\section{Conclusions}
We have presented a parametric DR method by training neural networks. In addition to the generalization of reducing the dimensionality of unseen data, the method achieved competitive performance and embedding quality compared to non-parametric state-of-the-art methods. Through utilizing the gradient clipping strategy, we trained the networks by directly optimizing the objective function from scratch instead of a pre-computed embedding. Furthermore, mini-batch sampling greatly reduces computation cost because the conditional probabilities of low-dimensional data points in each iteration are determined based on only batch samples. These two simple, yet effective, strategies make our parametric method a powerful and practical DR system for both general and streaming data. We will release our codes for public use after the paper is accepted for publication.

\bibliographystyle{aaai}
\bibliography{references}

\section{Appendix}
\subsection{Objective Function of LargeVis}
In LargeVis, the symmetric joint probability $p_{ij}$ is interpreted as the weight of an edge $e_{ij}$ that connects $x_i$ and $x_j$ in a high dimensional KNN graph; and $p_{ij}=0$ implies that edge $e_{ij}$ does not exist. Similarly, the unnormalized joint probability $w_{ij}$ (Eq. \ref{eq:tsne_Q}) indicates the probability of an edge between data points $y_i$ and $y_j$ in low-dimensional embedding space. Accordingly, the likelihood of an edge with weight $x$ in the embedding space can be defined as follows:
\begin{equation}
    P(e_{ij} = x) = P(e_{ij}=1)^x = w_{ij}^{p_{ij}}.
\end{equation}
Let $G_d$ and $G_e$ be the KNN graphs in high- and low-dimensional spaces, respectively. LargeVis attempts to maximize the likelihood of $G_e = G_d$
\begin{equation}
    L = \prod_{(i, j) \in E}w_{ij}^{p_{ij}} \prod_{(i, j) \in \overline{E}} (1-w_{ij})^{\gamma},
\end{equation}
where $\gamma$ is the weight assigned to negative edges. The objective function can then be rewritten as minimizing the negative log likelihood of $G_e = G_d$:
\begin{equation}
    C = -[\sum_{(i, j) \in E}p_{ij} log(w_{ij}) + \sum_{(i, j) \in \overline{E}} \gamma log(1-w_{ij})].
    \label{eq:largevis-loss}
\end{equation}

The optimization of Eq. (\ref{eq:largevis-loss}) is computationally expensive because the process has to consider all of the edges. To improve performance, we applied the negative sampling strategy to sample a positive edge according to $p_{ij}$ and five negative edges randomly to form a sample. In addition, since the positive edge was chosen according to its probability, we set equal weights to edges in the sample. We then used the batch of samples to update the network by minimizing the negative log-likelihood. This sampling process was precomputed for network optimization. Let $C^+$ and $C^-$ be the mappings of positive and negative edges, we derive the gradients:
\begin{align}
    \nonumber
    \frac{\partial C^-}{\partial \theta} & = \sum_{(i, j) \in \overline{E}} \frac{2 \gamma}{(\epsilon+d_{ij}^2)(1+d_{ij}^2)}(f(x_i) - f(x_j)) \frac{\partial f_\theta(x_i)}{\partial \theta},
    \\
    \frac{\partial C^+}{\partial \theta} & = \sum_{(i, j) \in E} \frac{-2}{1+d_{ij}^2}(f(x_i) - f(x_j)) \frac{\partial f_\theta(x_i)}{\partial \theta}, 
    \label{eq:nn-largevis-gradient}
\end{align}
where $d_{ij} = ||f(x_i)- f(x_j)||_2$ and $\epsilon$ is a small constant.

\subsection{Objective Function of UMAP}
UMAP models the similarity of data by using fuzzy sets. A fuzzy set $A = (U, m)$ is a pair, where $U$ is a set, and $m: U \to [0, 1]$ is a membership function that determines if an element in $U$ is a member of $A$. UMAP defines the fuzzy set $V_i$ = $(X, m_i^d)$ to represent if $x_j$ is a neighbor of $x_i$ in high-dimensional space, where $X$ is the set of all data points and
\begin{equation}
    m_i^d(x_i, x_j) = v_{j|i} = exp[-(r_{ij}-\rho_{i}) / \sigma_{i}],
\end{equation}
$r_{ij}$ is the distance between $x_i$ and $x_j$; $\rho_i$ is the distance to the nearest neighbor of $x_i$; and $\sigma_i$ works as the same perplexity calibration in t-SNE. Similarly, in the embedding space, UMAP defines a fuzzy set $U_i = (Y, m_i^e)$ with $Y$ as the set of all of the corresponding points and
\begin{equation}
    m_i^e(y_i, y_j) = w_{ij} = \frac{1}{1+a \cdot d_{ij}^b},
\end{equation}
where $d_{ij}$ is the distance between $y_i$ and $y_j$; $a$ and $b$ are hyperparameters that control the tightness of the squashing function. To find the optimized embeddings, UMAP minimizes the cross entropy between the fuzzy sets $U_i$ and $V_i$:
\begin{equation}
    C = \sum_{ij} [v_{ij}log(\frac{v_{ij}}{w_{ij}})+(1-v_{ij})log(\frac{1-v_{ij}}{1-w_{ij}})].
\end{equation}
In our implementation, we treated $v_{ij}$ the same as $p_{ij}$ in Eq. (\ref{eq:largevis-loss}) and applied the same sampling method to train the neural network. Similarly, we derive the gradients for the positive and negative edges as
\begin{align}
    \nonumber
    \frac{\partial C^-}{\partial \theta} = & \sum_{(i, j) \in \overline{E}} \frac{b}{(\epsilon+d_{ij}^2)(1+d_{ij}^2)}(f(x_i) - f(x_j)) \frac{\partial f_\theta(x_i)}{\partial \theta},
    \\
    \frac{\partial C^+}{\partial \theta} = & \sum_{(i, j) \in E} \frac{-2abd_{ij}^{2(b-1)}}{1+d_{ij}^2}(f(x_i) - f(x_j)) \frac{\partial f_\theta(x_i)}{\partial \theta},
    \label{eq:nn-largevis-gradient}
\end{align}
where $d_{ij} = ||f(x_i)- f(x_j)||_2$, $a$ and $b$ are hyper-parameters, and $\epsilon$ is a small constant.



\end{document}